\documentclass{article}

\usepackage{zhmanu}
\usepackage{graphicx,amsmath,amsfonts}
\allowdisplaybreaks%
\usepackage{xcolor}
\definecolor{codegray}{rgb}{0.5,0.5,0.5}
\usepackage{listings}
\newcommand{\ones}{{\bf 1{}}}  
\newcommand{\reals}{{\bf R{}}}  
\newcommand{\ints}{{\bf Z{}}}  
\newcommand{\prob}{\mathop{\bf prob}}  
\newcommand{\argmax}{\mathop{\rm argmax}}  
\newcommand{\norm}[1]{{\lVert #1 \rVert}}  
\newcommand{\inv}[1]{{#1}^{-1}}  

\newcommand{\eg}{{\it e.g.}}
\newcommand{\ie}{{\it i.e.}}
\newcommand{\etc}{{\it etc.}}
\newcommand{\etal}{{\it et al.}}

\newcommand{\states}{{\cal S{}}}
\newcommand{\actions}{{\cal A{}}}

\usepackage{lipsum}

\usepackage[colorlinks]{hyperref}

\title{Inverse Reinforcement Learning via\\Convex Optimization}
\author{Hao Zhu}
\author{Yuan Zhang}
\author{Joschka Boedecker}
\affil{Department of Computer Science and IMBIT//BrainLinks-BrainTools\\University of Freiburg}

\begin{document}
\maketitle

\begin{abstract}
    We consider the inverse reinforcement learning (IRL) problem, where an unknown reward function of some Markov decision process is estimated based on observed expert demonstrations.
    In most existing approaches, IRL is formulated and solved as a nonconvex optimization problem, posing challenges in scenarios where robustness and reproducibility are critical.
    We discuss a convex formulation of the IRL problem (CIRL) initially proposed by Ng and Russel, and reformulate the problem such that the domain-specific language \texttt{CVXPY} can be applied directly to specify and solve the convex problem.
    We also extend the CIRL problem to scenarios where the expert policy is not given analytically but by trajectory as state-action pairs, which can be strongly inconsistent with optimality, by augmenting some of the constraints.
    Theoretical analysis and practical implementation for hyperparameter auto-selection are introduced.
    This note helps the users to easily apply CIRL for their problems, without background knowledge on convex optimization.
\end{abstract}

\section{Introduction}
Inverse reinforcement learning (IRL) problems~\cite{ng2000algorithms} aim to estimate an unknown reward function of some Markov decision process (MDP), based on observed expert demonstrations.
A variety of IRL problem formulations have been proposed over the years, and have been widely used in engineering such as autonomous driving and robotics (see Arora and Doshi~\cite{arora2021survey}, Ruiz-Serra and Harr{\'e}~\cite{ruiz2023inverse} for a detailed review).
However, most of these formulations are nonconvex and thus can only be approximately solved by searching for a local optimum, which is normally achieved via some variants of subgradient methods.
As a result, convergence is not always guaranteed in these methods and usually a relatively large number of iterations are required to find the (approximate) solution.
Moreover, such approximate solutions can lead to issues in some specific application domains where robustness and reproducibility are highly required.
In psychology and neuroscience research, for instance, IRL appears to be emerging as mathematical behavior models for estimating the intrinsic reward function underlying animal or human subjects' behavior.
Since a large number of subsequent analyses and experiment designs might be dependent on the IRL problem solution, lack of reproducibility can cause severe problems~\cite{yamaguchi2018identification,kwon2020inverse,alyahyay2023mechanisms}.
To promote the application of IRL in such areas, in this note, we consider the \textit{convex IRL} (CIRL) problem formulation proposed by Ng and Russel~\cite{ng2000algorithms}, addressing two major limitations of their original work:
First, while the authors have provided theoretical analysis and some numerical experiments demonstrating the feasibility of CIRL, the unmentioned solver implementation can be cumbersome and error-prone.
Since many practitioners are not well versed in this procedure, the use of this method is limited to convex optimization experts.
Second, the originally proposed CIRL would be infeasible when the expert behavior is strongly inconsistent with optimality, \eg, when the expert alternates between multiple subgoals that conflict with each other.
This is one common case when the expert demonstrations are represented as a sequence of state-action pairs in the environment.
Moreover, we also provide theoretical analysis and practical implementation for automatically selecting some of the hyperparameters of CIRL.
All code related to this note is open-sourced under:
\begin{center}
   \url{https://github.com/nrgrp/cvx_irl}.
\end{center}

The rest of this note is arranged as follows:
Background knowledge about MDPs and our basic notation is introduced in \S\ref{sec:mdp}.
In \S\ref{sec:cirl}, we briefly review the CIRL problem formulation from Ng and Russel~\cite{ng2000algorithms} and extend some theoretical details about hyperparameter selection.
We then provide a reformulation of the original CIRL problem such that it can be typed into some domain specific languages (DSLs) for convex optimization (\S\ref{sec:prob_reform}) such as \texttt{CVXPY}~\cite{diamond2016cvxpy}, and extend it to estimating multiple reward functions corresponding to different subgoals of the expert (\S\ref{sec:subgoal}).
Implementation of a solver using \texttt{CVXPY}, and some application examples, are provided in \S\ref{sec:impl} and \S\ref{sec:examp}, respectively.

\section{Convex inverse reinforcement learning}
\subsection{Markov decision processes}\label{sec:mdp}
We consider a finite MDP given by a tuple $(\states,\ \actions,\ {\{P_a\}}_{a \in \actions},\ r,\ \gamma)$.
The sets $\states$ and $\actions$ are both finite sets, consisting of all possible states and actions of the MDP, respectively.
In the rest of this note, we let $|\states| = m$ and $|\actions| = k$.
The set ${\{P_a\}}_{a \in \actions}$ consists of all transition probability matrices $P_a \in \reals^{m \times m}$ upon taking action $a \in \actions$, \ie, each entry of the matrix $P_{ij, a} = \prob(s_j \mid s_i, a)$, $i, j = 1, \ldots, m$.
Obviously, $P_a$ has to satisfy $P_a \ones = \ones$, \ie, the sum of each row of $P_a$ equals to $1$.
The function $r \colon \states \to \reals$ defines the reward function, which is assumed to be unknown under the context of IRL.
Note that we write $r$ as a function only on the set $\states$ for simplicity, while it can be readily extended to be on $\states \times \actions$.
Besides, we also assume the reward function is bounded by some positive real number $r^{\rm max} \in \reals_{++}$.
For a finite MDP, we can overload the reward function defined above as a vector $r \in \reals^m$, and we will use this vector representation of the reward function in the rest of this note.
(Although we will still refer to it as the reward `function'.)
The scalar $\gamma \in [0, 1)$ is the discount factor.

\subsection{CIRL problem formulation}\label{sec:cirl}
\paragraph{Policy and optimality condition.}
A (deterministic) policy for an MDP is defined as a function $\pi \colon \states \to \actions$.
Suppose the observed expert policy $\pi$ is optimal for an MDP with some unknown reward function $r$, to simplify the notation, we can assume $\pi(s) = a^\star \in \actions$ for all $s \in \states$ without losing of generality.
(This can be done by reordering the actions and permuting the rows between all $P_a$ if necessary.)
The corresponding state transition matrix is then $P_{a^\star}$, and the value function (in vector form) $v \in \reals^m$ evaluated for this policy $\pi$ is
\begin{equation}\label{eq:value_fn}
    v = r + \gamma P_{a^\star} v \implies v = \inv{(I - \gamma P_{a^\star})} r,
\end{equation}
where the matrix $I - \gamma P_{a^\star}$ is always invertible as $\gamma < 1$~\cite{sutton2018reinforcement}.
The optimality of this policy $\pi$ suggests that the corresponding reward function $r$ must satisfy the Bellman optimality condition~\cite{sutton2018reinforcement}
\begin{align}
    &\pi(s_i) = a^\star \in \argmax_{a \in \actions} p_{i, a}^T v,\quad i = 1, \ldots, m\notag\\
    \Longleftrightarrow\quad & P_{a^\star} v \succeq P_a v,\quad \mbox{for all $a \in \actions\setminus  \{a^\star\}$}\notag\\
    \Longleftrightarrow\quad & P_{a^\star} \inv{(I - \gamma P_{a^\star})} r \succeq P_a \inv{(I - \gamma P_{a^\star})} r,\quad \mbox{for all $a \in \actions\setminus  \{a^\star\}$}\notag\\
    \Longleftrightarrow\quad & (P_{a^\star} - P_a) \inv{(I - \gamma P_{a^\star})} r \succeq 0,\quad \mbox{for all $a \in \actions\setminus  \{a^\star\}$},\label{eq:opt_cond}
\end{align}
where $p_{i, a}^T$ denotes the $i$th row of $P_a$, and $\succeq$ denotes componentwise inequality between vectors.

\paragraph{The CIRL problem.}
The optimality condition (\ref{eq:opt_cond}) and boundedness assumption in \S\ref{sec:mdp} provide a solution set of the reward functions for the IRL problem.
However, such a set contains some trivial solutions, \eg, $r = c$, where $c \in \reals^m$, $c_1 = \cdots = c_m$ is any constant vector.
To choose a nontrivial (or `meaningful') reward function from this solution set, we consider the following optimization problem:
\begin{equation}\label{prob:cirl_gen}
    \begin{array}{ll}
        \mbox{minimize} & J(r) + \lambda \phi(r)\\
        \mbox{subject to} & (P_{a^\star} - P_a) \inv{(I - \gamma P_{a^\star})} r \succeq 0,\quad \mbox{for all $a \in \actions\setminus  \{a^\star\}$}\\
        & r^{\rm max} \succeq r \succeq -r^{\rm max},
    \end{array}
\end{equation}
where the unknown reward function $r \in \reals^m$ is the variable, ${\{P_a\}}_{a \in \actions}$ and $\gamma$ are the problem data, and $r^{\rm max} > 0$, $\lambda \geq 0$ are the hyperparameters.
Note that we use the aforementioned solution set as the feasible set of (\ref{prob:cirl_gen}).
(The boundedness constraint on $r$ can also be tightened to be asymmetric as, \eg, $r^{\rm max} \succeq r \succeq 0$, if some prior information about $r$ such as nonnegativity is given.
Nevertheless, we will keep the general case as in (\ref{prob:cirl_gen}) for the rest of this note.)
The two functions $J(r)$ and $\phi(r)$ represent two criteria that a reward function is considered to be meaningful~\cite{ng2000algorithms}.
Firstly, it is natural to favor reward functions that maximize the margin between the observed expert policy $\pi$ and all other possible policies at all states as much as possible, corresponding to the primary objective:
\begin{align}
    J(r) &= -\sum_{i = 1}^m \left(p_{i, a^\star}^Tv - \sup_{a \in \actions\setminus \{a^\star\}} p_{i, a}^Tv\right)\notag\\
    &= -\sum_{i = 1}^m \inf_{a \in \actions \setminus \{a^\star\}} (p_{i, a^\star}^T - p_{i, a}^T) v\notag\\
    &= -\sum_{i = 1}^m \inf_{a \in \actions \setminus \{a^\star\}} \left((p_{i, a^\star}^T - p_{i, a}^T) \inv{(I - \gamma P_{a^\star})} r\right)\notag,
\end{align}
where $p_{i, a}^T$ denotes the $i$th row of $P_a$, and the last equality is from substituting (\ref{eq:value_fn}).
Secondly, it is also believed that the reward function $r$ should be as sparse as possible, resulting in the $\ell_1$-penalty function $\phi(r) = \norm{r}_1$.
Put together, we have
\begin{equation}\label{prob:cirl_sc}
    \begin{array}{ll}
        \mbox{minimize} & -\sum_{i = 1}^m \inf_{a \in \actions \setminus \{a^\star\}} \left((p_{i, a^\star}^T - p_{i, a}^T) \inv{(I - \gamma P_{a^\star})} r\right) + \lambda \norm{r}_1\\
        \mbox{subject to} & (P_{a^\star} - P_a) \inv{(I - \gamma P_{a^\star})} r \succeq 0,\quad \mbox{for all $a \in \actions\setminus \{a^\star\}$}\\
        & r^{\rm max} \succeq r \succeq -r^{\rm max},
    \end{array}
\end{equation}
which is a convex optimization problem originally proposed by Ng and Russel~\cite{ng2000algorithms} for IRL problems.
The convexity of problem (\ref{prob:cirl_sc}) is readily shown:
The two constraints are all linear inequality constraints on $r$, and the objective is convex since it is a nonnegative sum of a sequence of convex functions, \ie, the negative infimum over affine $-\inf_{a \in \actions \setminus \{a^\star\}} \left((p_{i, a^\star}^T - p_{i, a}^T) \inv{(I - \gamma P_{a^\star})} r\right)$, $i = 1, \ldots, m$, and the $\ell_1$-norm function $\norm{r}_1$, which are all convex functions about variable $r$~\cite{boyd2004convex}.

\subsection{Hyperparameter selection}
The scalarization weight $\lambda$ is a free parameter such that by varying it in $[0, \infty)$ we trade off between the primary objective $J(r)$ and the penalty term $\phi(r)$.
Specifically, there exists a value
\begin{equation}\label{eq:lbdmax}
    \lambda^{\rm max} = \inf_{z \in \partial J(0)} \norm{z}_\infty,
\end{equation}
where $\partial J(0)$ is the subdifferential of $J(r)$ at $r = 0$, such that if $\lambda \geq \lambda^{\rm max}$, the optimal of (\ref{prob:cirl_gen}) is achieved at $r = 0$.
To show this, note that $r = 0$ is always in the feasible set defined by the constraints of (\ref{prob:cirl_gen}), thus to ensure the optimal value is achieved by $r = 0$ for a given $\lambda$, we only need to guarantee
\begin{equation*}
        0 \in \partial J(0) + \lambda \partial \norm{x}_1 \Big|_{x = 0}\quad
        \Longleftrightarrow\quad -\partial J(0) \cap \lambda \partial \norm{x}_1 \Big|_{x = 0} \neq \emptyset.
\end{equation*}
This is equivalent to requiring that there exists some $z \in \partial J(0)$ such that
\begin{equation*}
    -z \in \lambda \partial \norm{x}_1 \Big|_{x = 0}\quad \Longleftrightarrow\quad \norm{z}_\infty \leq \lambda,
\end{equation*}
which is guaranteed if $\lambda \geq \inf_{z \in \partial J(0)} \norm{z}_\infty$.
This corresponds to the critical value
\begin{equation*}
    \lambda^{\rm max} = \inf_{z \in \partial J(0)} \norm{z}_\infty
\end{equation*}
as stated in (\ref{eq:lbdmax}), which ends the proof.

By calculating $\lambda^{\rm max}$ according to (\ref{eq:lbdmax}) and assigning the value to $\lambda$, one can find the `simplest' $r$ under the given problem data.
In practice, however, instead of directly calculating $\lambda^{\rm max}$, it is more common to use iterative methods, such as bisection, to find a $\lambda < \lambda^{\rm max}$ that is close to $\lambda^{\rm max}$ under some threshold.
(We demonstrate an implementation of this procedure for automatically determining $\lambda$ in \S\ref{sec:impl_autotune}.)
Regarding the reward function bound $r^{\rm max}$, since it has no influence on the indices of the nonzero entries of $r$, but just tight the absolute value of these nonzero entries such that (\ref{prob:cirl_gen}) is not unbounded below, one can simply assign their favorite positive number to $r^{\rm max}$ when specifying the problem.

\subsection{CIRL in regularized linear program form}\label{sec:prob_reform}
We introduce a reformulation of the problem (\ref{prob:cirl_sc}) such that it can be directly specified and solved by \texttt{CVXPY}.
Introducing the epigraph variable $s \in \reals^m$ for the primary objective
\begin{equation*}
    J(r) = -\sum_{i = 1}^m \inf_{a \in \actions \setminus \{a^\star\}} \left((p_{i, a^\star}^T - p_{i, a}^T) \inv{(I - \gamma P_{a^\star})} r\right),
\end{equation*}
the problem (\ref{prob:cirl_sc}) is straightforwardly equivalent to
\begin{equation}\label{prob:reglp}
    \begin{array}{ll}
        \mbox{minimize} & \ones^T s + \lambda \norm{r}_1\\[5pt]
        \mbox{subject to} & \left[
            \begin{array}{c}
                p_{i, a^\star}^T - p_{i, \tilde{a}_1}^T\\
                \vdots\\
                p_{i, a^\star}^T - p_{i, \tilde{a}_{k-1}}^T\\
            \end{array}
            \right]
            \inv{(I - \gamma P_{a^\star})} r + s_i \succeq 0,\quad i = 1, \ldots, m\\[25pt]
        &(P_{a^\star} - P_{\tilde{a}_i}) \inv{(I - \gamma P_{a^\star})} r \succeq 0,\quad i = 1, \ldots, k-1\\[5pt]
        & r^{\rm max} \succeq r \succeq -r^{\rm max},
    \end{array}
\end{equation}
where $\tilde{a}_1, \ldots, \tilde{a}_{k-1} \in \actions\setminus \{a^\star\}$ (which is trivially equivalent to $a_1, \ldots, a_{k-1}$ if $a^\star = a_k$).
The problem (\ref{prob:reglp}) is then an $\ell_1$-regularized linear program over variables $s$ and $r$, with $2mk$ inequality constraints.
Notice that the optimal value of (\ref{prob:reglp}) is expected to be attained with some $s \preceq 0$, given that the problem is feasible.
Thus one can also explicitly introduce $s \preceq 0$ as an additional constraint on $s$ to (\ref{prob:reglp}), but it will not influence the solution.
In practice, the potential prior sparsity pattern of the involved matrices in (\ref{prob:reglp}) can be readily incorporated to accelerate the matrix operations.

\subsection{Reward estimation for subgoals}\label{sec:subgoal}
The optimality condition (\ref{eq:opt_cond}) requires that the observed expert policy is globally optimal under the unknown reward function.
However, in real-world applications, this is not necessarily the case, especially when the expert's policy is not directly given but represented as a sequence of state-action pairs.
For instance, in many behavioral experiments, \eg, Hamaguchi \etal~\cite{hamaguchi2022prospective}, De La Crompe \etal~\cite{de2023freibox}, the subject needs to achieve a series of subgoals such that the whole task is fulfilled, \ie, the globally optimal policy consists of a series of locally optimal policies (and might even conflict with each other).
In these scenarios, directly solving (\ref{prob:reglp}) is likely to have the trivial solution $r = 0$ returned, which is useless to the researchers.

To make the solution of (\ref{prob:reglp}) useful when the given expert demonstration is only locally optimal (optimal within a subset of all states during some specific time period), we introduce relaxation on some of the constraints below.
Consider the expert policy is indirectly represented by a sequence of $n$ state-action pairs ${\cal D} = ((s_1, a_1), \ldots, (s_n, a_n))$, where $s_1, \ldots, s_n \in \states$ and $a_1, \ldots, a_n \in \actions$.
Let the $n$-demonstrations partitioned by indices $t_0, \ldots, t_p \in \ints_{++}$, $t_0 = 1 < t_1 < \cdots < t_p = n$, and each of the partition ${\cal D}_j = ((s_{t_j}, a_{t_j}),\ \ldots,\ (s_{t_{j+1}}, a_{t_{j+1}}))$, $j = 0, \ldots, p-1$, corresponds to a locally optimal policy within the subset of states $\states_j = \{s \mid (s, a) \in \{(s_{t_j}, a_{t_j}),\ \ldots,\ (s_{t_{j+1}}, a_{t_{j+1}})\}\}$.
(In practice, such partition indices $t_0, \ldots, t_p$ can be obtained using different methods, such as via change point detection of time series~\cite{jia2024chainofthought}, or by fitting some latent variable models~\cite{zhu2024multiintention}, \etc, which is, however, not the major focus of this note.)
We can then relax the first and second constraints in (\ref{prob:reglp}) from global optimality on the set of all states $\states$ to local optimality on the subset of visited states $\states_j$ under demonstrations in ${\cal D}_j$, $j = 0, \ldots, p$, \ie,
\begin{equation}\label{eq:relax_margin}
    \left[
            \begin{array}{c}
                \tilde{p}_{i, a^\star}^T - \tilde{p}_{i, \tilde{a}_1}^T\\
                \vdots\\
                \tilde{p}_{i, a^\star}^T - \tilde{p}_{i, \tilde{a}_{k-1}}^T\\
            \end{array}
            \right]
            \inv{(I - \gamma \tilde{P}_{a^\star})} r + s_i \succeq 0,\quad i = 1, \ldots, |\states_j|
\end{equation}
and
\begin{equation}\label{eq:relax_optimcond}
    (\tilde{P}_{a^\star} - \tilde{P}_{\tilde{a}_i}) \inv{(I - \gamma \tilde{P}_{a^\star})} r \succeq 0,\quad i = 1, \ldots, k-1,
\end{equation}
where the matrices $\tilde{P}_{a^\star}, \tilde{P}_{\tilde{a}_1}, \ldots, \tilde{P}_{\tilde{a}_{k-1}} \in \reals^{|\states_j| \times |\states_j|}$ are the submatrices of $P_{a^\star}, P_{\tilde{a}_1}, \ldots, P_{\tilde{a}_{k-1}}$, respectively, by only taking those entries related to the states in $\states_j$, and the vector $\tilde{p}_{i, a}^T \in \reals^{|\states_j|}$ are again the $i$th row of matrix $\tilde{P}_a$.
The objective as well as the box constraint on $r$ of (\ref{prob:reglp}) remain unchanged.
By solving this relaxed problem of (\ref{prob:reglp}) for expert policy given by ${\cal D}_1, \ldots, {\cal D}_p$ we can obtain the reward function $r_j$ corresponding to the $j$th subgoal, $j = 1, \ldots, p$.
One advantage of such relaxation is that the implementation for specifying and solving the problem (\ref{prob:reglp}) (see \S\ref{sec:impl}) can be directly applied without further adaptation.

A small technical condition we should note is that some rows of the submatrices $\tilde{P}_a$ in (\ref{eq:relax_margin}) and (\ref{eq:relax_optimcond}) might no longer satisfy $\tilde{p}^T_{i, a} \ones = 1$.
Under many cases, this violation will only introduce some approximation error, but it can also cause severe issues when the matrix $I - \gamma \tilde{P}_{a^\star}$ is not invertible.
We address this problem by renormalizing each row of $\tilde{P}_a$ if necessary as follows.
If $\tilde{p}^T_{i, a} \neq 0$, we can simply overload $\tilde{p}^T_{i, a}$ with the normalized version $\tilde{p}^T_{i, a}/(\tilde{p}^T_{i, a} \ones)$, $i = 1, \ldots, |\states_j|$, to guarantee it is still a proper probability distribution.
If $\tilde{p}^T_{i, a} = 0$ (which can happen if the transition matrix $P_a$ is deterministic), we explicitly set the $i$th entry of $\tilde{p}^T_{i, a}$ equals $1$, \ie, $(\tilde{p}^T_{i, a})_i = 1$.
Such operation can be interpreted as introducing a `barrier' around state $s_i$ under action $a$, where performing the action will simply make the state unchanged.

We introduce an example involving the above procedure in \S\ref{sec:examp_subgoal}.

\section{Implementation}\label{sec:impl}
We depict below the \texttt{CVXPY} code that specifies and solves (\ref{prob:reglp}).

\noindent\hrulefill%
\begin{lstlisting}[language=Python]
import numpy as np
import cvxpy as cp

# problem information (input from user)
m = None  # number of states
gamma = None  # discount factor
Pastr = None  # transition matrix of the optimal action
lPa = []  # list of transition matrices of the other actions

# hyperparameters (input from user)
rmax = None  # reward function bound
lbd = None  # scalarization weight

r = cp.Variable(m)
s = cp.Variable(m)

constraints = []
H = np.linalg.inv(np.identity(m) - gamma * Pastr)
D = np.array([[Pastr[i] - Pa[i] for Pa in lPa] for i in range(m)])
for i in range(m):
    constraints.append(D[i] @ H @ r + s[i] >= 0)
for Pa in lPa:
    constraints.append((Pastr - Pa) @ H @ r >= 0)
constraints.append(rmax >= r)
constraints.append(r >= -rmax)

obj = cp.Minimize(cp.sum(s) + lbd * cp.norm(r, 1))
prob = cp.Problem(obj, constraints)
prob.solve()
\end{lstlisting}
\hrulefill%

To fully specify the problem, users only have to assign the required problem information according to their MDP and expert policy, as well as the hyperparameters $r^{\rm max}$ and $\lambda$.
Note that in this implementation $\lambda$ is declared as a constant value.
If one would like to hand-tune the value of $\lambda$, especially when $m$ is large, declaring $\lambda$ as a \texttt{Parameter} object of \texttt{CVXPY} would substantially save the computing time upon solving the problem multiple times with different $\lambda$ values (see \S\ref{sec:impl_autotune}).

\section{Examples}\label{sec:examp}
In this section, we show two examples of IRL via solving the convex optimization problem (\ref{prob:reglp}).
All experiments were performed on an AMD Ryzen\texttrademark\ 9 7950X (4.5 GHz) CPU.
The code to reproduce these examples is available at
\begin{center}
    \url{https://github.com/nrgrp/cvx_irl}.
\end{center}

\subsection{Example 1: Gridworld}
\begin{figure}[t]
    \centering
    \includegraphics[width=0.8\textwidth]{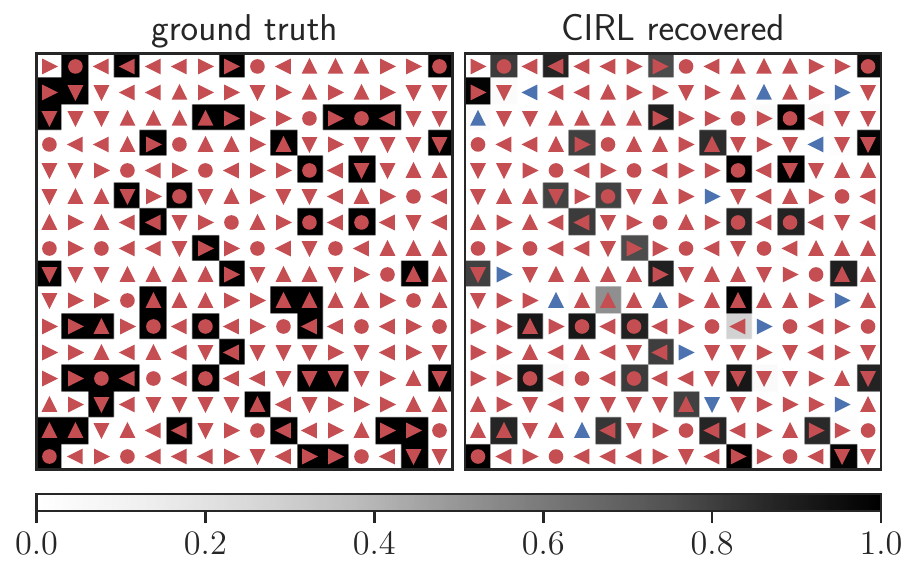}
    \caption{
        \textit{Left.}
        The $16 \times 16$ gridworld.
        The grey colormap depicts the reward distribution in the environment.
        The red arrows (and dots for the action \textit{stay}) indicate the optimal moving direction under the aforementioned reward distribution.
        \textit{Right.}
        Recovered reward distribution and corresponding optimal policy from CIRL.
        The inconsistent optimal policy compared to the ground truth is colored blue.
        Both the ground truth and recovered reward function are normalized to be in the range of $[0, 1]$.
    }\label{fig:gw}
\end{figure}

In the first example we consider a $16 \times 16$ gridworld, where the agent can choose to go \textit{left}, \textit{right}, \textit{up}, \textit{down}, or \textit{stay} at the current state.
Upon each action execution, except for moving towards the selected direction, there will always be a $10\%$ chance of moving randomly out of all $5$ possible directions.
The ground truth reward distribution and corresponding optimal policy are shown in the left part of figure~\ref{fig:gw}.

To recover the true reward function given the depicted optimal policy, the problem (\ref{prob:reglp}) was solved with hyperparameter $\lambda = 2$ and $r^{\rm max} = 100$.
The recovered reward function and the corresponding optimal policy under this recovered reward are shown in the right part of figure~\ref{fig:gw}.
It can be seen immediately that the recovered reward function aligns well with the true reward.
Quantitatively, the cosine similarity between the true reward and the CRIL recovered reward is $0.85$, where this similarity between the true and some random reward should be around $0$.
Besides, by comparing the true optimal policy and the optimal policy under the recovered reward, the fraction of matched action over all states is 0.94.
Note that in this example, the running time for solving (\ref{prob:reglp}) without incorporating any matrix sparsity pattern is $1.65$ seconds, which is to the best of our knowledge significantly faster than the other existing IRL algorithms, whose computing time are generally measured in minutes under an environment with similar scale as a $16 \times 16$ gridworld.

\subsection{Example 2: The greedy snake}\label{sec:examp_subgoal}

\begin{figure}[p]
    \centering
    \includegraphics[width=0.8\textwidth]{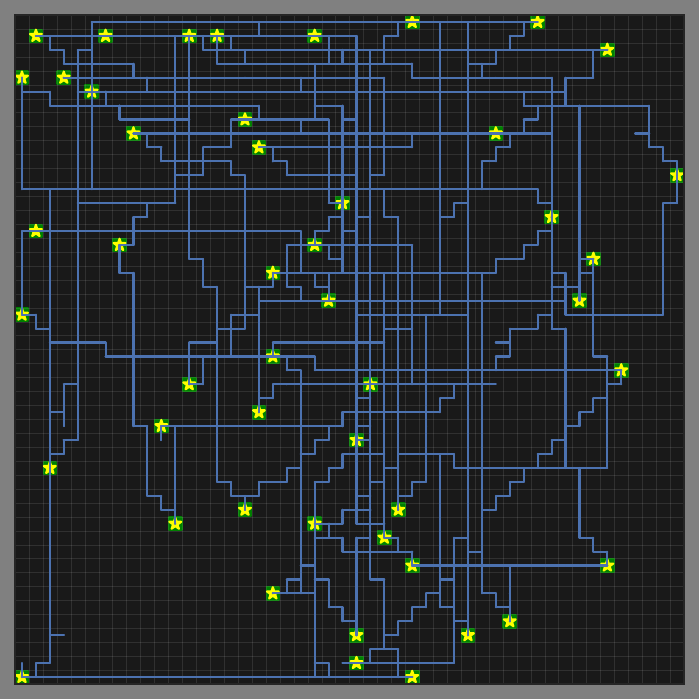}
    \caption{
        The greedy snake environment.
        The blue line shows the expert's trajectory.
        The ground truth and estimated reward distribution for each trajectory segmentation are depicted as green squares and yellow stars, respectively.
    }\label{fig:sn}
\end{figure}

We consider a $48 \times 48$ gridworld environment with nonstationary reward distribution.
Only one state is rewarded at each time.
An expert (the greedy snake) will chase for the reward, and once the reward is collected, another reward will be generated at a random state.
This task is a typical example where the expert needs to alternate between multiple subgoals, such that its policy might conflict with each other throughout the whole episode, but is locally (in time) optimal for each reward location.
The action space consists of $5$ actions: \textit{left}, \textit{right}, \textit{up}, \textit{down}, and \textit{stay}.
Upon each action execution of the expert, except for moving towards the selected direction, there will always be a $10\%$ chance of moving randomly out of all $5$ possible directions.
The expert's policy is given by its trajectory as a sequence of state-action pairs ${\cal D}$, and the partition time where the rewarded state changed is also given.
(In practice, such partition time might indeed not be known, but various methods can be applied to obtain this information according to the application scenarios~\cite{jia2024chainofthought,zhu2024multiintention}.
Here we assume the trajectory partition is given only for the convenience of demonstrating the feasibility of estimating the nonstationary reward function according to the expert's trajectory via solving (\ref{prob:reglp}) with augmented constraints (\ref{eq:relax_margin}) and (\ref{eq:relax_optimcond}).)
The hyperparameters are set to be $r^{\rm max} = 100$ and $\lambda = 0.5$ when solving (\ref{prob:reglp}) for all trajectory partitions.

Figure~\ref{fig:sn} shows the ground truth and estimated reward location for individual trajectory segments.
The estimated reward locations were obtained by taking the state with the largest estimated reward value, for individual trajectory segments.
The full expert trajectory consists of $2022$ actions with $48$ reward location changes.
It can be seen that our recovered reward distribution matches exactly to the ground truth.
This leads to a $0.94$ fraction of the matched action through the whole trajectory between the observed expert action and the predicted optimal action under the estimated reward.
We also provide a video showing the aforementioned results in figure~\ref{fig:sn}, which can be found at
\begin{center}
    \url{https://github.com/nrgrp/cvx_irl}.
\end{center}
The average computing time for the individual trajectory segments are $9.01 \pm 1.10$ ms (mean $\pm$ standard error), which corresponds to a total computing time of $432$ ms.

These results indicate that by relaxing the constraints of problem (\ref{prob:reglp}) to (\ref{eq:relax_margin}) and (\ref{eq:relax_optimcond}), we are able to apply CIRL for nonstationary reward function without knowing the expert's policy at all states.
In the original work from Ng and Russel~\cite{ng2000algorithms}, if the expert's policy is given by trajectories, Monte Carlo simulation is required to obtain the full analytical expert policy $\pi$, which can be time-consuming even under moderate scaled environments, and their required optimality condition for all states will make the problem (\ref{prob:reglp}) infeasible (or only feasible at trivial solution $r = 0$) where the expert's policy is only locally optimal for some time period before the reward function changing.

\section{Related work and comments}
\paragraph{Inverse reinforcement learning.}
The IRL problem was initially proposed by Ng and Russel~\cite{ng2000algorithms}.
Many variants of the IRL problem formulations has been introduced over the decades, such as maximum margin methods~\cite{abbeel2004apprenticeship}, probabilistic methods~\cite{lopes2009active}, and the most widely used maximum entropy methods~\cite{ziebart2008maximum}.
(The references listed here only serve as examples for different types of methods for IRL; interested readers can refer to Arora and Doshi~\cite{arora2021survey}, Ruiz-Serra and Harr{\'e}~\cite{ruiz2023inverse} for a detailed comprehensive review.)
To the best of our knowledge, except for the method from Ng and Russel~\cite{ng2000algorithms} mentioned in this note, most of these IRL problem formulations are nonconvex.
Exceptions are some maximum margin methods, in particular, which are based on linear programming (\eg, Syed \etal~\cite{syed2008apprenticeship}).
These methods, however, focus more on apprenticeship learning, which aims at obtaining a policy $\pi$ comparable with the experts, instead of the reward function itself.
Hence, the reward function $r$ is only a byproduct of the primary objective in these methods and sometimes is not even returned explicitly.
In this note, on the contrary, we consider the IRL formulation where the primary objective is to obtain the unknown reward function $r$ according to which the expert optimized their policy.
Besides, we do not tend to depreciate the class of maximum entropy IRL methods which has been widely used in different areas, instead, we only claim that the CIRL method could be more suitable for the scenarios where reproducibility and robustness of the IRL results are highly required, or the running time is strictly constrained.

\paragraph{Contribution.}
Our major contributions in extending the original work by Ng and Russel~\cite{ng2000algorithms} are threefold.
First, we reformulated the CIRL problem into the epigraph form such that it can be easily typed into some DSLs for convex optimization.
In particular, we provide an implementation of CIRL based on \texttt{CVXPY}, which allows practitioners that are not well versed in convex optimization to have access to applying this method easily.
Second, we augment the constraints of CIRL such that it can be applied in scenarios where the expert's behavior is given by trajectory as state-action pairs, which can be strongly inconsistent with optimality.
One common instance is when the expert alternates between multiple subgoals that conflict with each other, as demonstrated in \S\ref{sec:examp_subgoal}.
Last but not least, we provide a thorough theoretical analysis of the selection of the $\ell_1$-penalty parameter $\lambda$, which is (conceptually) useful if one would like to obtain the most sparse reward function given some problem data without obtaining the trivial solution $r = 0$.
As a complementary for the theory, we also introduce a \texttt{CVXPY} implementation for fast automatic tuning of the hyperparameter $\lambda$ in practice.

\section*{Acknowledgements}
This work has been funded as part of BrainLinks-BrainTools, which is funded by the Federal Ministry of Economics, Science and Arts of Baden-Württemberg within the sustainability program for projects of the Excellence Initiative II, and CRC/TRR 384 `IN-Code'.

\bibliography{refs}

\newpage
\appendix
\setcounter{equation}{0}
\numberwithin{equation}{section}
\section{Implementation of auto-tuning the scalarization weight}\label{sec:impl_autotune}
In this section, we extend the implementation of CIRL in \S\ref{sec:impl} to automatically select the hyperparameter $\lambda$ to find the reward function $r$ with maximum sparsity given the problem data, which can make full use of the capacity of \texttt{CVXPY}.

\noindent\hrulefill%
\begin{lstlisting}[language=Python]
import numpy as np
import cvxpy as cp

# problem information (input from user)
m = None  # number of states
gamma = None  # discount factor
Pastr = None  # transition matrix of the optimal action
lPa = []  # list of transition matrices of the other actions

# hyperparameters (input from user)
rmax = None  # reward function bound
lbd_up = None  # upper bound on lambda
lbd_low = None  # lower bound on lambda
epsilon = None  # terminate if lbd_up - lbd_low <= epsilon

r = cp.Variable(m)
s = cp.Variable(m)
lbd = cp.Parameter(nonneg=True)

constraints = []
H = np.linalg.inv(np.identity(m) - gamma * Pastr)
D = np.array([[Pastr[i] - Pa[i] for Pa in lPa] for i in range(m)])
for i in range(m):
    constraints.append(D[i] @ H @ r + s[i] >= 0)
for Pa in lPa:
    constraints.append((Pastr - Pa) @ H @ r >= 0)
constraints.append(rmax >= r)
constraints.append(r >= 0)

obj = cp.Minimize(cp.sum(s) + lbd * cp.norm(r, 1))
prob = cp.Problem(obj, constraints)

while True:
    lbd.value = 0.5 * (lbd_up + lbd_low)
    opt_val = prob.solve()

    if np.abs(opt_val) < 1e-6:
        lbd_up = lbd.value
    elif lbd_up - lbd_low <= epsilon:
        break
    else:
        lbd_low = lbd.value
\end{lstlisting}
\hrulefill%

In this implementation, the hyperparameter $\lambda$ is declared as a nonnegative instance of the object \texttt{CVXPY.Parameter} instead of simply a number, such that the computing time can be saved when trying to re-solve the optimization problem multiple times using different $\lambda$ values during the auto-tuning process.
The user needs to specify a range on $\lambda$ so that the bisection process will be automatically performed based on this interval.
Once the width of the interval of $\lambda$ is below some threshold \texttt{epsilon} given by the user, and the optimal value of (\ref{prob:reglp}) is not $0$, the algorithm will quit with the current $\lambda$ and problem solution as the output.

The code listed above can be found at
\begin{center}
    \url{https://github.com/nrgrp/cvx_irl}.
\end{center}

\end{document}